\documentclass[letterpaper,journal]{IEEEtran}
\usepackage{amsmath,amssymb,bm}
\usepackage{graphicx}
\usepackage{booktabs}
\usepackage{multirow} % for merged table cells
\usepackage[table]{xcolor}
\usepackage{textcomp}
\usepackage{url}
\usepackage{cite}
\usepackage[font=footnotesize,labelfont=bf]{caption}
\usepackage{subcaption}
\usepackage{cuted} % provides non-floating full-width blocks in two-column layout

\newcommand{\ours}{\textbf{Ours}}
\newcommand{\group}[2]{\multicolumn{#1}{l}{\emph{#2}} \\}

\title{\LARGE \bf Facet-SLAM: Real-Time Triangle Splatting SLAM}

\author{
    Edward Beng Wai Tan$^{1,*}$, Siew-Kei Lam$^{1}$%
    \thanks{$^{1}$College of Computing and Data Science, Nanyang Technological University, Singapore.}%
    \thanks{$^{*}$Corresponding author.}%
}

\begin{document}
\maketitle
\thispagestyle{empty}\pagestyle{empty}

\begin{strip}
    \vspace{-2.5\baselineskip}
    \centering
    \includegraphics[width=\textwidth]{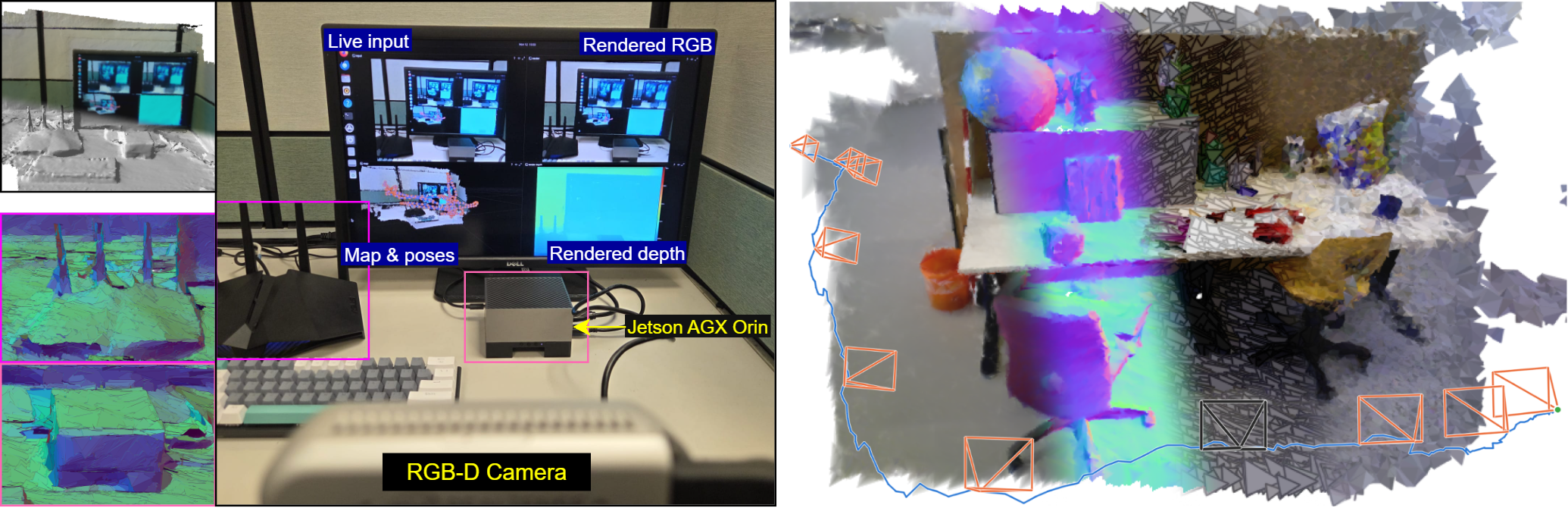}

    \captionof{figure}{(\textbf{Right}) Our method produces a photorealistic triangle map in real-time, and with accurate geometry and enables mesh generation. (\textbf{Left}) The same system is pictured running live on a Jetson AGX Orin at sensor rate. To the extreme left are the processed TSDF and surface normals of the observed scene. 
    }
    \label{fig:hook}
\end{strip}
\thispagestyle{empty}
\pagestyle{empty}

\begin{abstract}
Dense SLAM systems based on 3D Gaussian splatting build maps that render photorealistically, but their surface geometry is poor, and obtaining a mesh requires extraction or fusion stages. Triangle splatting avoids this with surface primitives whose vertices can be triangulated directly into a connected mesh, yet SLAM with this representation has so far relied on photometric tracking, which is too slow for localization at sensor rate. We present Facet-SLAM, the first real-time triangle splatting SLAM system for RGB-D cameras. Fast geometric registration is the natural alternative, but it is not sufficient alone, since the mapper's optimization displaces triangles from the observed surface and makes them unreliable registration targets. We therefore separate the two roles: the tracker sets each triangle's centroid from fused depth and registers frames to these centroids by GICP, while the mapper is restricted to non-translating updates via a change of variables. Our system achieves competitive tracking accuracy at around 100 FPS on a desktop GPU and runs at sensor rate on a Jetson AGX Orin. Compared with the photometric approach, it is over 20 times faster, matches rendering quality, and improves surface accuracy, reducing the chamfer distance of the Delaunay mesh on Replica from 1.55 to 0.86\,cm. We will release the implementation at https://github.com/edward62740/Facet-SLAM.
\end{abstract}

\section{Introduction}
Recent SLAM systems use differentiable representations such as 3D Gaussian splatting (3DGS) \cite{kerbl3DGaussianSplatting2023} to build maps that render photorealistically. This is useful for applications such as visual navigation and embodied AI training, but many of these also need geometrically accurate maps. 3DGS maps often have poor surface geometry, and obtaining a good mesh requires post-hoc extraction \cite{guedonSuGaRSurfaceAlignedGaussian2023}, an intermediate representation \cite{guedonMILoMeshIntheLoopGaussian2025,yuGaussianOpacityFields2024}, or 2DGS \cite{huang2DGaussianSplatting2024} primitives requiring fusion. In contrast, the primitives of a differentiable triangle soup \cite{Held2025Triangle} are themselves surface elements, so a connected mesh can be obtained by triangulating their existing vertices, without an intermediate representation or a separate fusion stage.

More recently, Triangle Splatting SLAM \cite{fry2026triangle} (TS-SLAM) introduced the first dense SLAM method based on triangle primitives, demonstrating the above-mentioned geometric properties, in addition to the ability to generate a photorealistic map. However, TS-SLAM lacks real-time performance, limiting deployment flexibility and its use for on-device localization while constructing the map. 

We present, to the best of our knowledge, the first \textit{real-time} triangle splatting SLAM method. It runs in real-time on both desktop GPU and Jetson AGX Orin, with pose accuracy competitive with photometric methods, rendering quality on par with TS-SLAM, and more accurate map geometry than prior triangle splatting and 3DGS SLAM methods.

Our contributions are summarized as follows:
\begin{itemize}
    \item We introduce a real-time triangle splatting SLAM system that runs at sensor rate on a Jetson AGX Orin, and from which a connected mesh can be extracted directly.
    \item We propose a tracking-and-mapping strategy that makes triangles reliable GICP registration targets, by fixing their centroids to fused depth and restricting the mapper to non-translating updates.
    \item We improve surface accuracy over both prior triangle splatting and 3DGS SLAM methods while maintaining visual quality relative to triangle splatting methods.
\end{itemize}

\begin{figure*}[t]
\centering\includegraphics[width=\textwidth]{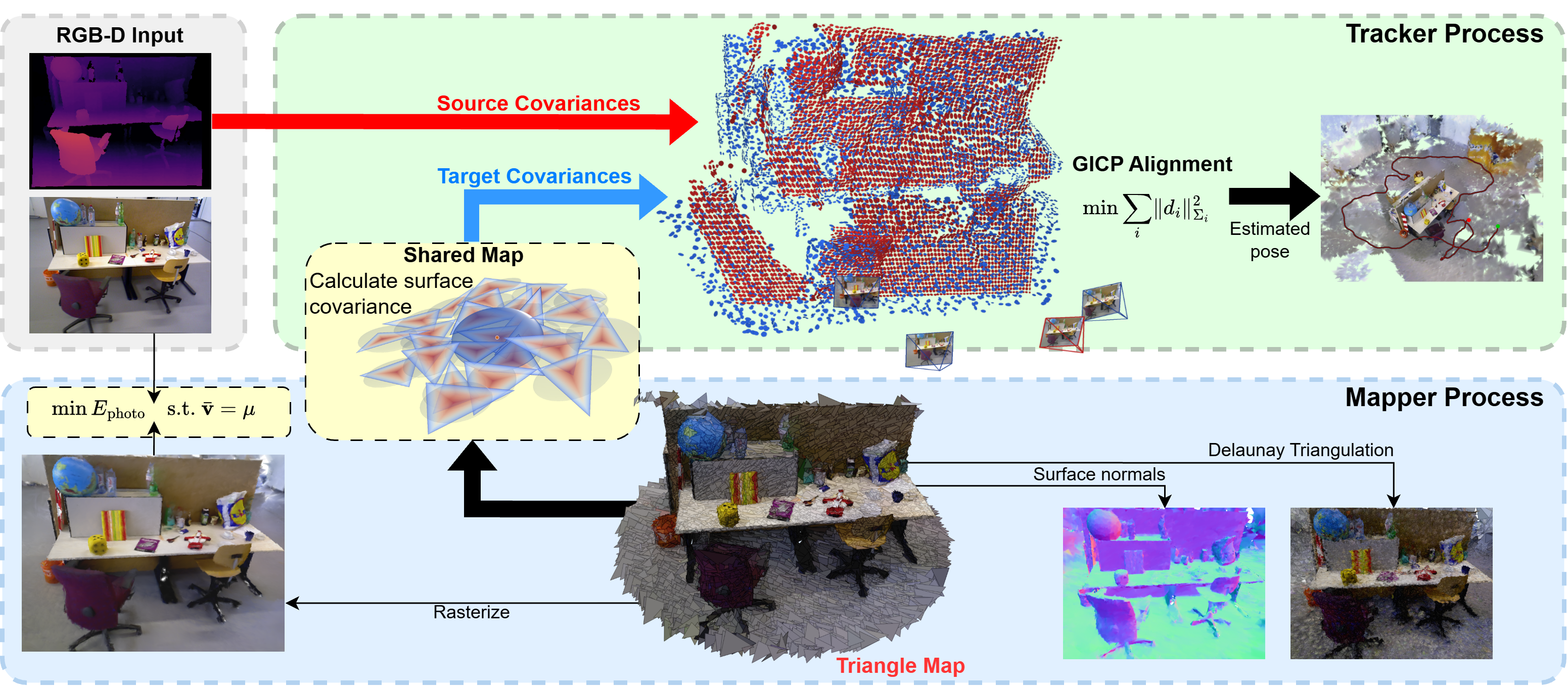}
\caption{\textbf{System overview.} The system runs as two processes. The mapper (GPU) optimizes a
triangle map by rendering, subject to the mean of each triangle's vertices $\bar{\mathbf{v}}$ staying
at the position $\boldsymbol{\mu}$ measured by depth, and publishes the triangle means to the tracker. The
tracker (CPU) registers every depth frame to these means by GICP, with the local scatter of the depth
points and of the neighbouring means as covariances, and returns keyframes with their poses to the
mapper.}
\label{fig:system}
\end{figure*}

\section{Related Work}

\textbf{Photorealistic SLAM} methods, such as 3DGS SLAM, initially estimated pose by minimizing dense photometric error \cite{matsukiGaussianSplattingSLAM2024, yugayGaussianSLAMPhotorealisticDense2024, keethaSplaTAMSplatTrack2024}, which is accurate but too slow for real-time localization. Later works reach real-time operation with faster trackers, such as Photo-SLAM \cite{huangPhotoSLAMRealtimeSimultaneous2024} which builds on ORB-SLAM3 \cite{camposORBSLAM3AccurateOpenSource2021a}, and GS-ICP SLAM \cite{haRGBDGSICPSLAM2025} which uses GICP \cite{segalGeneralizedICP2009}.

\textbf{Surface geometry} remains a limitation of 3DGS maps. Obtaining a good mesh from 3DGS requires post-hoc extraction \cite{guedonSuGaRSurfaceAlignedGaussian2023}, an intermediate representation \cite{guedonMILoMeshIntheLoopGaussian2025,yuGaussianOpacityFields2024}, or surface-aligned primitives followed by volumetric fusion \cite{huang2DGaussianSplatting2024}; the latter have been adopted into SLAM system \cite{pakG2SICPSLAMGeometryaware2025}. In contrast, the primitives of a differentiable triangle soup \cite{Held2025Triangle} are themselves surface elements, so a connected mesh can be obtained by triangulating their existing vertices, without an intermediate representation or a separate fusion stage.

\textbf{Triangle splatting} \cite{Held2025Triangle} was recently proposed as an alternative to 3DGS, and TS-SLAM \cite{fry2026triangle} is the first dense SLAM method to use triangles as map primitives, motivated by the need for accurate surfaces in embodied AI applications. Like the early 3DGS SLAM methods, TS-SLAM estimates pose photometrically, which is accurate but runs at only a few frames per second, too slow for on-device localization.

\textbf{Point cloud registration} against the map has been used in 3DGS SLAM by reusing the Gaussian properties such as the covariances \cite{haRGBDGSICPSLAM2025}, or means \cite{tanDualCovarianceGaussian2026}. A triangle appears to be a better registration target still, being a plane with an explicit normal. However, we show that point-to-plane registration on the triangles' own normals is unreliable on real depth, which we attribute to the mapper setting each triangle's orientation to minimize a composite mapper cost. We therefore take the GICP covariance from the scatter of neighboring triangle centroids, and fix those centroids to fused depth by restricting the map optimization to be non-translating.

\section{Method}

Our system (see Fig.~\ref{fig:system}) performs tracking and mapping in parallel \cite{klein07parallel}, where the tracker uses GICP \cite{segalGeneralizedICP2009} and the mapper uses triangle primitives \cite{fry2026triangle,Held2025Triangle} to model the scene. Each triangle $j$ has vertices $\mathbf{v}_{j1},\mathbf{v}_{j2},\mathbf{v}_{j3}\in\mathbb{R}^3$, each with a color $\mathbf{c}\in\mathbb{R}^3$ and opacity $o\in[0,1]$. Triangle centroids $\boldsymbol{\mu}_j$ are set by the tracker from depth observations, while the mapper optimizes the remaining degrees of freedom.

\subsection{Tracking}
\label{sec:cov}

GICP~\cite{segalGeneralizedICP2009} registers observation points $\mathbf{a}_i$ to map points $\mathbf{b}_i$, each pair modeled as draws from $\mathcal{N}(\mathbf{a}_i^*,\mathbf{A}_i)$ and $\mathcal{N}(\mathbf{b}_i^*,\mathbf{B}_i)$ with $\mathbf{b}_i^*=\mathbf{T}\mathbf{a}_i^*$:
\begin{equation}
\mathbf{T}^*=\arg\min_{\mathbf{T}}\sum_i \mathbf{d}_i^\top\big(\mathbf{B}_i+\mathbf{R}\mathbf{A}_i\mathbf{R}^\top\big)^{-1}\mathbf{d}_i,\quad \mathbf{d}_i=\mathbf{b}_i-\mathbf{T}\mathbf{a}_i,
\label{eq:gicp}
\end{equation}
to determine the MLE $\mathbf{T}^*\in\mathrm{SE}(3)$. A triangle's own geometry is a poor choice for the target covariance $\mathbf{B}_i$, since its vertices are supervised only indirectly through the rendering loss (see Section~\ref{subsec:ablation}). Instead, we approximate the local surface distribution at each triangle's centroid by the scatter of its $K$ nearest neighbouring centroids $\mathcal{K}_i$, plus an isotropic noise floor
\begin{equation}
\mathbf{B}_i=\frac{1}{K}\sum_{l\in\mathcal{K}_i}(\boldsymbol{\mu}_l-\bar{\boldsymbol{\mu}}_i)(\boldsymbol{\mu}_l-\bar{\boldsymbol{\mu}}_i)^\top+\sigma_f^2 \mathbf{I} .
\label{eq:map}
\end{equation}

The observation covariances $\mathbf{A}_i$ are likewise computed from the local scatter of depth points. Since triangle centroids are associated with tracking keyframes and are not translated by the mapper, we additionally perform pose-only window optimization over the latest $N$ keyframes, with geometric and photometric residuals between overlapping keyframes, and move the associated triangles rigidly with the correction. This gives a marginal improvement in ATE on longer sequences.

\subsection{Mapping}
\label{sec:map}
Triangles are rasterized as in \cite{Held2025Triangle}, with $\phi_j(\mathbf{p})$ the signed distance of pixel $\mathbf{p}$ to the projected edges and $\mathbf{s}_j$ the incenter, the window $w_j(\mathbf{p})=\max\big(0,\phi_j(\mathbf{p})/\phi_j(\mathbf{s}_j)\big)^{\sigma}$ falls from one at the incenter to zero at the edges, with sharpness $\sigma$. Pixel color is alpha-composited in depth order, $\hat{I}(\mathbf{p})=\sum_j \mathbf{c}_j(\mathbf{p})\,o_jw_j(\mathbf{p})\prod_{i<j}\big(1-o_iw_i(\mathbf{p})\big)$, with $\mathbf{c}_j(\mathbf{p})$ being the barycentric interpolation of the vertex colors and $o_j$ the mean vertex opacity.

The map is optimized with a composite loss:
\begin{equation}
E_{\mathrm{photo}}=E_{\mathrm{rgb}}+\lambda_dE_{\mathrm{dep}}+\lambda_nE_{\mathrm{norm}}+\lambda_eE_{\mathrm{reg}},
\label{eq:emap}
\end{equation}
in which surface position is supervised only indirectly as a sum of terms. Noise and bias in the photometric gradients therefore displaces triangles from the observed surface, even on flat regions, and GICP registration against them degrades.

We address this by a separation of duties: the tracker sets each triangle's centroid as the running mean of the keyframe depth observations associated with it, while the mapper optimizes appearance, shape and orientation.

On the mapper side, we enforce the fixed centroid by a linear change of variables. For each coordinate axis, let $\mathbf{x}\in\mathbb{R}^3$ stack that coordinate of the three vertices and $\mu$ be the corresponding centroid coordinate. We define
\begin{equation}
\mathbf{x}=\mu\mathbf{1}+\mathbf{Q}\mathbf{e},\qquad \mathbf{e}\in\mathbb{R}^2,
\label{eq:pin}
\end{equation}
where the columns of $\mathbf{Q}\in\mathbb{R}^{3\times2}$ are an orthonormal basis of the plane orthogonal to $\mathbf{1}$.\footnote{Any such basis satisfies the constraint; we use $(1,-1,0)^\top/\sqrt{2}$ and $(1,1,-2)^\top/\sqrt{6}$ in our implementation.} Adam optimizes $\mathbf{e}$ through $\nabla_{\mathbf{e}}E_{\mathrm{photo}}=\mathbf{Q}^\top\nabla_{\mathbf{x}}E_{\mathrm{photo}}$, which is the vertex gradient with its translational component removed. The loss itself is unchanged from Eq.~\ref{eq:emap}.

\section{Experiments}
\subsection{Setup}
\label{sec:setup}
\textbf{Datasets and Metrics.} We evaluate on TUM RGB-D~\cite{sturmBenchmarkEvaluationRGBD2012}, Replica~\cite{straubReplicaDatasetDigital2019} and ScanNet \cite{daiScanNetRichlyAnnotated3D2017}. We report RMSE ATE (Absolute Trajectory Error) for pose estimation, PSNR, SSIM \cite{zhouwangImageQualityAssessment2004}, LPIPS \cite{zhangUnreasonableEffectivenessDeep2018} for 3D reconstruction quality. We also report L1 (depth) error, chamfer, accuracy and surface completeness to characterize the geometric quality of the map. We \textbf{bold} the best and \underline{underline} the second best results.

\textbf{Implementation details.} We run the experiments on two hardware platforms, a desktop computer with RTX 4090 GPU and i9-13900KF CPU, and a Jetson AGX Orin (MAXN mode) embedded platform. All baselines are reported with either their reported results, or our reproduced numbers where possible. We implemented reproductions for TS-SLAM \cite{fry2026triangle}, Photo-SLAM \cite{huangPhotoSLAMRealtimeSimultaneous2024}, GS-ICP SLAM \cite{haRGBDGSICPSLAM2025} and MonoGS \cite{matsukiGaussianSplattingSLAM2024}. We also compare with SplaTAM \cite{keethaSplaTAMSplatTrack2024} and GS-SLAM\cite{yanGSSLAMDenseVisual2024}. For fairness and due to the stochasticity of some methods, we run all measurements 5 times per sequence and report the average.

\subsection{Tracking}
In Table~\ref{tab:tum} we show our method achieves competitive tracking on the 3 standard TUM RGB-D scenes, performing similarly to the dense photometric methods in 3DGS, such as MonoGS and SplaTAM, and also marginally below TS-SLAM, while having significantly higher FPS on both desktop GPU and Jetson AGX Orin. Our method also performs comparably to real-time 3DGS-based SLAM methods (e.g. Photo-SLAM, GS-ICP SLAM), while still being faster. Similar to GS-ICP SLAM, we report the ATE capped at 30 FPS (camera rate), though there is little impact on the ATE.

\begin{table}[h]
\centering\caption{ATE [cm] $\downarrow$ on TUM RGB-D scenes, and average FPS.}
\label{tab:tum}
\footnotesize\setlength{\tabcolsep}{2.5pt}
\begin{tabular}{lcccc | c}
\toprule
\textbf{Method} & \textbf{fr1/desk} & \textbf{fr2/xyz} & \textbf{fr3/office} & \textbf{Avg} & \textbf{FPS} $\uparrow$ \\
\midrule
\group{6}{3D Gaussian Splatting}
SplaTAM & 3.4 & 1.2 & 5.2 & 3.3 & 0.4 \\
GS-SLAM & 3.3 & 1.3 & 6.6 & 3.7 & -- \\
MonoGS & \textbf{1.44} & 1.77 & \underline{1.61} &{1.61} & 2.5 \\
GS-ICP SLAM & 3.94 & 2.28 & 3.26 & 3.16 & \underline{86} \\
Photo-SLAM & 2.30 & \textbf{0.36} & \textbf{0.96} & \textbf{1.21} & 40 \\
\midrule
\group{6}{Triangle Splatting}
TS-SLAM& \underline{1.77} & 1.12 & 1.83 & \underline{1.57} & 1.5 \\
\ours & 2.13 & \underline{1.08} & 2.07 & 1.76 & \textbf{113} \\
\ours\ (Jetson) & 2.26 & 1.26 & 1.87 & 1.79 & 36 \\
\bottomrule
\end{tabular}\\
\end{table}

We report more extensive results on TUM RGB-D scenes in Table~\ref{tab:tumext}, showing that our method maintains parity with other methods. In addition, our method's ATE on Jetson is largely preserved despite the lower average FPS and other computational bottlenecks, demonstrating the robustness of our tracking method to less frequent keyframes (as the GPU thread is slower). We also report on more diverse scenes, including the Replica dataset in Table~\ref{tab:replica}, consisting of simulated indoor scenes, and on ScanNet dataset in Table~\ref{tab:scannet}. In both datasets, our method performs similarly to 3DGS and triangle splatting baselines. 

\begin{table}[h]
\centering\caption{ATE [cm] $\downarrow$ on additional TUM RGB-D scenes, and average FPS.}
\label{tab:tumext}
\footnotesize\setlength{\tabcolsep}{2.5pt}
\resizebox{\columnwidth}{!}{%
\begin{tabular}{lcccccccc}
\toprule
\textbf{Method} & \textbf{desk2} & \textbf{xyz} & \textbf{360} & \textbf{rpy} & \textbf{st\_far} & \textbf{st\_near} & \textbf{Avg} & \textbf{FPS} $\uparrow$ \\
\midrule
\group{9}{3D Gaussian Splatting}
MonoGS & 5.75 & \textbf{0.92} & \textbf{7.14} & 2.07 & \textbf{0.89} & \underline{1.44} & \textbf{3.04} & 2.2 \\
GS-ICP SLAM  & 15.58 & 1.57 & 45.45 & 4.63 & 3.43 & 4.78 & 12.57 & \underline{60} \\
Photo-SLAM  & \underline{2.49} & 0.98 & 15.66 & \underline{2.01} & \underline{1.02} & \textbf{0.96} & 3.85 & 44 \\
\midrule
\group{9}{Triangle Splatting}
TS-SLAM  & \textbf{2.40} & \textbf{0.92} & 13.00 & \textbf{1.40} & 1.39 & 1.60 & \underline{3.45} & 3.9 \\
\ours & 3.66 & 1.35 & 10.69 & 2.51 & 1.52 & 2.81 & 3.76 & \textbf{94} \\
\ours\ (Jetson) & 4.08 & 1.55 & \underline{10.22} & 2.26 & 1.86 & 2.14 & 3.69 & 27 \\
\bottomrule
\end{tabular}}
\end{table}

\begin{table}[h]
\centering\caption{ATE [cm] $\downarrow$ on ScanNet.}
\label{tab:scannet}
\footnotesize\setlength{\tabcolsep}{2.5pt}
\begin{tabular}{lccccccc}
\toprule
\textbf{Method} & \textbf{0000} & \textbf{0059} & \textbf{0106} & \textbf{0169} & \textbf{0181} & \textbf{0207} & \textbf{Avg} \\
\midrule
\group{8}{3D Gaussian Splatting}
MonoGS & 22.6 & 11.1 & 9.4 & \underline{10.3} & 28.7 & 8.8 & \textbf{15.2} \\
GS-ICP SLAM & 31.2 & 30.2 & 10.3 & 29.8 & \textbf{20.4} & 10.8 & 22.1 \\
Photo-SLAM & \textbf{8.2} & \underline{6.9} & \underline{8.9} & \textbf{8.3} & 80.0 & \textbf{7.1} & 19.9 \\
\midrule
\group{8}{Triangle Splatting}
TS-SLAM & \underline{21.9} & \textbf{6.1} & 9.4 & 61.1 & \underline{21.8} & 10.8 & 21.8 \\
\ours & 26.1 & 14.5 & \textbf{7.6} & 33.1 & 28.1 & \underline{7.6} & \underline{19.5} \\
\bottomrule
\end{tabular}
\end{table}

\begin{figure*}[t]
\centering\includegraphics[width=\textwidth]{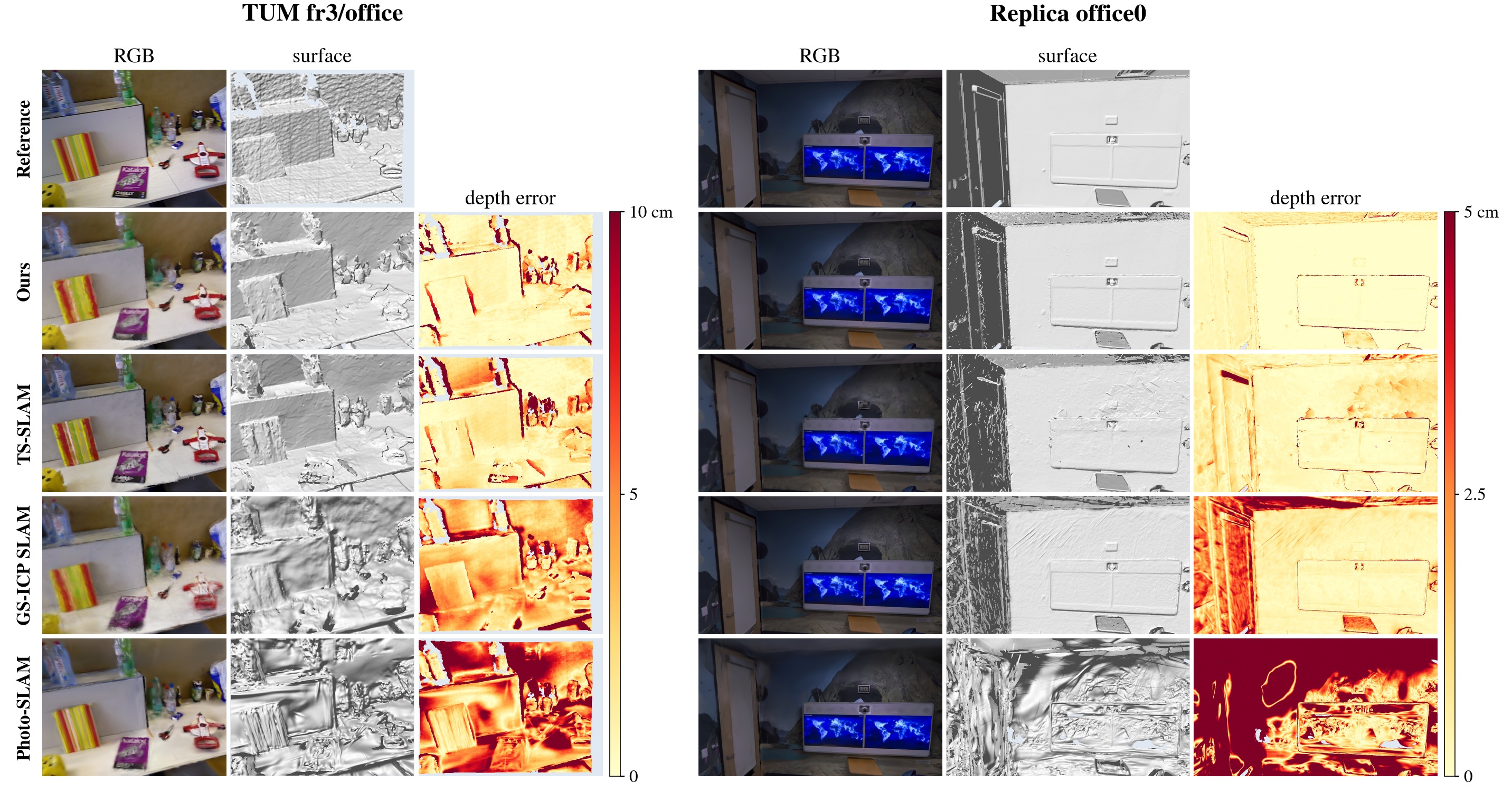}
\caption{\textbf{Qualitative renders of surface error.} Both ours and TS-SLAM have substantially better depth error and less visible unevenness than the GS-based methods, and ours has slightly better surface quality than TS-SLAM despite having worse RGB rendering quality.}
\label{fig:geom}
\end{figure*}

\begin{table}[t]
\centering\caption{ATE [cm] $\downarrow$ on Replica.}
\label{tab:replica}
\setlength{\tabcolsep}{1.8pt}
\begin{tabular}{lccccccccc}
\toprule
\textbf{Method} & \textbf{r0} & \textbf{r1} & \textbf{r2} & \textbf{o0} & \textbf{o1} & \textbf{o2} & \textbf{o3} & \textbf{o4} & \textbf{Avg} \\
\midrule
\group{10}{3D Gaussian Splatting}
SplaTAM & \underline{0.31} & 0.40 & 0.29 & 0.47 & 0.27 & 0.29 & 0.32 & 0.55 & 0.36 \\
GS-SLAM & 0.48 & 0.53 & 0.33 & 0.52 & 0.41 & 0.59 & 0.46 & 0.70 & 0.50 \\
MonoGS & 0.49 & 0.26 & 0.35 & 0.48 & 0.49 & 0.22 & \underline{0.20} & 2.44 & 0.62 \\
GS-ICP SLAM & \textbf{0.16} & \textbf{0.17} & \underline{0.13} & \textbf{0.20} & \underline{0.13} & \underline{0.20} & 0.21 & \underline{0.22} & \underline{0.18} \\
Photo-SLAM & 0.51 & 0.38 & 0.27 & 0.50 & 0.45 & 1.24 & 0.81 & 0.60 & 0.60 \\
\midrule
\group{10}{Triangle Splatting}
TS-SLAM & 0.36 & 0.41 & 0.25 & 0.32 & 0.30 & 0.45 & 0.73 & 0.55 & 0.42 \\
\ours & \textbf{0.16} & \underline{0.19} & \textbf{0.11} & \underline{0.25} & \textbf{0.12} & \textbf{0.19} & \textbf{0.17} & \textbf{0.18} & \textbf{0.17} \\
\bottomrule
\end{tabular}\\
\end{table}

\subsection{Geometry}
We first report our results on the quality of the map's geometry in Table~\ref{tab:geom}. Our method outperforms both TS-SLAM, as well as the other 3DGS SLAM methods on the real-world TUM RGB-D dataset. We measure L1 depth error, chamfer distance, surface accuracy and completion within a 2cm threshold. These experiments demonstrate the quality of our method's maps when used for on-the-fly reconstruction, since these properties hold \textit{while} the SLAM system is in operation.

\begin{table}[h]
\centering\caption{Geometry on TUM RGB-D, depth L1 (cm) $\downarrow$, chamfer distance (cm) $\downarrow$, accuracy and completion within 2\,cm (\%) $\uparrow$.}
\label{tab:geom}
\scriptsize\setlength{\tabcolsep}{1.6pt}
\resizebox{\columnwidth}{!}{%
\begin{tabular}{lcccccccccccc}
\toprule
 & \multicolumn{3}{c}{\textbf{L1}} & \multicolumn{3}{c}{\textbf{chamfer}} & \multicolumn{3}{c}{\textbf{{acc.\,$<$2\,cm}}} & \multicolumn{3}{c}{\textbf{{comp.\,$<$2\,cm}}} \\
\cmidrule(lr){2-4}\cmidrule(lr){5-7}\cmidrule(lr){8-10}\cmidrule(lr){11-13}
\textbf{Method} & {desk} & xyz & office & desk & xyz & office & desk & xyz & office & desk & xyz & office \\
\midrule
\group{13}{3D Gaussian Splatting}
MonoGS & 6.67 & 15.26 & 14.78 & 5.27 & 9.81 & 7.88 & {31.6} & {18.1} & {24.8} & {44.6} & {37.8} & {35.4} \\
GS-ICP SLAM & 5.93 & 8.93 & 19.86 & 5.03 & 8.94 & 7.41 & {37.8} & {35.6} & {26.0} & {39.4} & {47.1} & {30.8} \\
Photo-SLAM & 10.24 & 52.34 & 29.78 & 9.81 & 24.12 & 15.88 & {22.6} & {7.0} & {23.2} & {28.6} & {26.2} & {36.5} \\
\midrule
\group{13}{Triangle Splatting}
TS-SLAM & 4.66 & 9.96 & 10.60 & 3.45 & \textbf{5.32} & 2.60 & {56.3} & {\textbf{60.9}} & {66.0} & {41.2} & {72.7} & {61.4} \\
\ours & \textbf{3.14} & \textbf{6.37} & \textbf{6.95} & \textbf{2.62} & 6.42 & \textbf{1.65} & {\textbf{64.9}} & {60.1} & {\textbf{77.4}} & {\textbf{53.0}} & {\textbf{74.9}} & {\textbf{70.3}} \\
\bottomrule
\end{tabular}}
\end{table}

\begin{table}[h]
\centering\caption{Geometry on Replica (avg.): depth L1 (cm) $\downarrow$, chamfer distance (cm) $\downarrow$ of the TSDF and Delaunay meshes, accuracy, completion and F-score within 1 or 2\,cm (\%) $\uparrow$.}
\label{tab:geomrep}
\scriptsize\setlength{\tabcolsep}{1.6pt}
\begin{tabular}{lcccccccccc}
\toprule
 & & \multicolumn{7}{c}{\textbf{TSDF}} & \multicolumn{2}{c}{\textbf{Delaunay}} \\
\cmidrule(lr){3-9}\cmidrule(lr){10-11}
\textbf{Method} & L1 & chamfer & {acc.$_1$} & {comp.$_1$} & {F$_1$} & {acc.$_2$} & {comp.$_2$} & {F$_2$} & chamfer & {F$_1$} \\
\midrule
\group{11}{3D Gaussian Splatting}
MonoGS & 4.93 & 3.38 & {17.1} & {24.7} & {20.1} & {39.9} & {59.0} & {47.4} & -- & {--} \\
GS-ICP SLAM & 2.50 & 1.94 & {19.9} & {20.8} & {20.3} & {66.0} & {70.9} & {68.4} & -- & {--} \\
Photo-SLAM & 17.40 & 14.90 & {5.5} & {8.4} & {6.7} & {13.2} & {24.6} & {17.0} & -- & {--} \\
\midrule
\group{11}{Triangle Splatting}
TS-SLAM & 0.68 & 0.95 & {73.5} & {76.1} & {74.8} & {91.6} & {94.4} & {93.0} & 1.55 & {54.1} \\
\ours & \textbf{0.65} & \textbf{0.67} & {\textbf{85.2}} & {\textbf{85.7}} & {\textbf{85.4}} & {\textbf{98.0}} & {\textbf{98.3}} & {\textbf{98.2}} & \textbf{0.86} & {\textbf{73.6}} \\
\bottomrule
\end{tabular}
\end{table}

Next, we evaluate the quality of the meshes extracted by TSDF fusion and restricted Delaunay triangulation following TS-SLAM's evaluation \cite{fry2026triangle} on Replica in Table~\ref{tab:geomrep}. We report chamfer distance (cm), and also accuracy and completeness @ \{1,2\} cm, with the corresponding F-score. Our method has the best performance on all metrics reported. We also report L1 depth errors on Replica prior to processing and omit Delaunay results on the 3DGS SLAM methods since there is no direct equivalent. We attribute a large part of our gains in geometry over TS-SLAM to our contributions in Section~\ref{sec:map}. We observe that although Photo-SLAM has highly accurate pose estimation, its geometric properties are relatively poor even compared to other 3DGS SLAM methods, which we attribute to the lack of explicit depth supervision after initialization. This can be empirically observed in Fig.~\ref{fig:geom}, where we provide a qualitative comparison of our method compared to TS-SLAM and the other real-time 3DGS SLAM methods. Despite having visibly worse RGB rendering quality in some cases, ours has the least visible errors in surface quality.

\subsection{Rendering}
\begin{figure}[h]
\centering\includegraphics[width=\columnwidth]{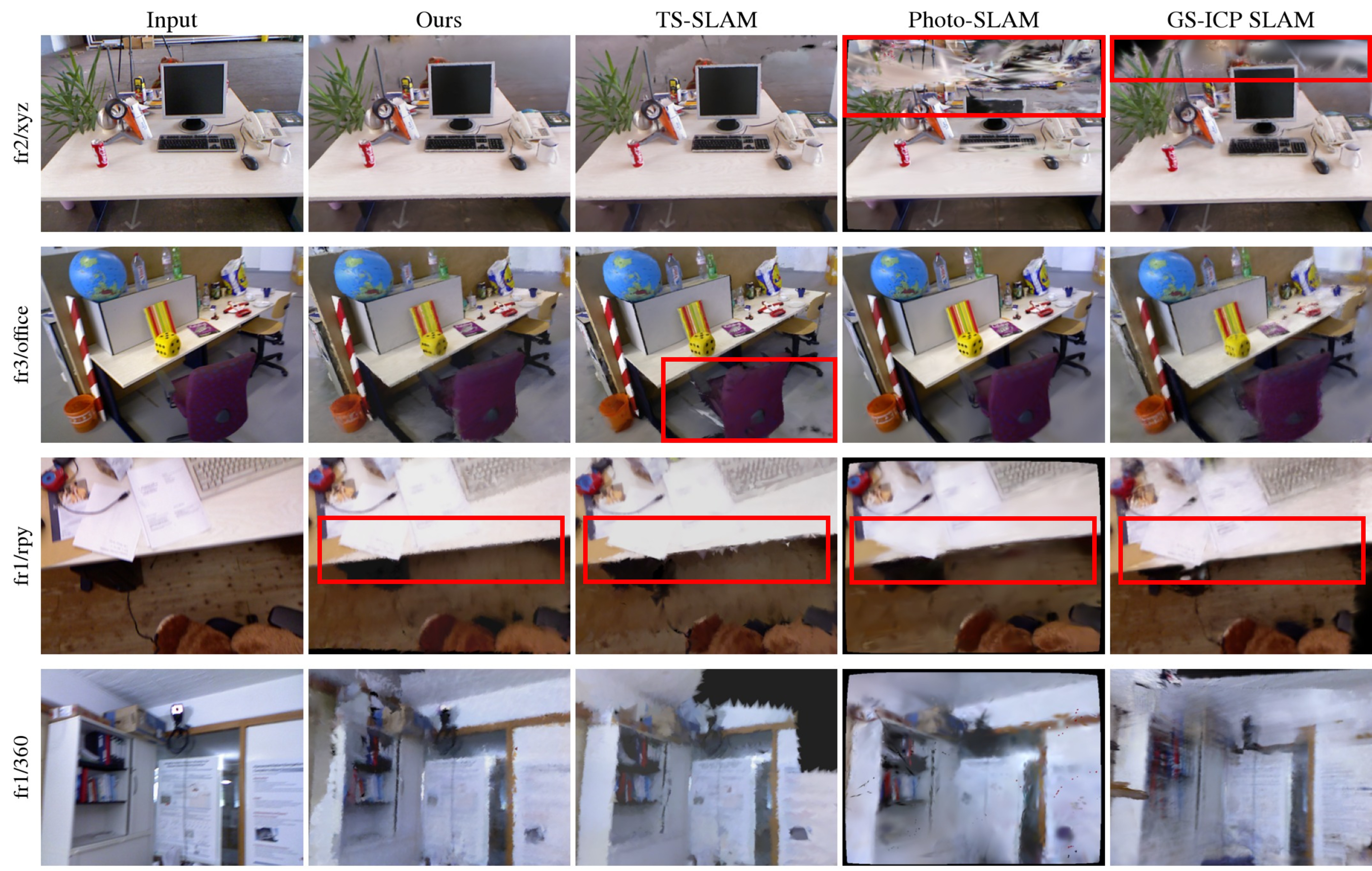}
\caption{\textbf{Qualitative rendering of test views.} Our method maintains good rendering quality over diverse scene types, including sharp details (xyz), shadows (office), sharp edges (rpy) and high camera rotation (360).}
\label{fig:render}
\end{figure}

We report our method's quantitative rendering quality on TUM RGB-D and ScanNet in Table~\ref{tab:render} and Table~\ref{tab:renderscn}. Similar to other triangle splatting methods, our method balances both realistic rendering and accurate geometry, and thus the map quality trails 3DGS methods. Note that part of this comparison is also dependent on certain baselines (e.g. Photo-SLAM) having higher-order SHs; we select 0 as a design decision, but this can be changed as a trade-off between computational and view-dependent render quality. We also provide a qualitative visualization in Fig.~\ref{fig:render}, where we select views that are non-training (novel) views for all the methods.

\begin{table}[h]
\centering\caption{Rendering quality on TUM RGB-D. }
\label{tab:render}
\footnotesize\setlength{\tabcolsep}{2.5pt}
\resizebox{\columnwidth}{!}{%
\begin{tabular}{llcccccccccc}
\toprule
\textbf{Method} & & \textbf{desk} & \textbf{fr2/xyz} & \textbf{office} & \textbf{desk2} & \textbf{xyz} & \textbf{360} & \textbf{rpy} & \textbf{st\_far} & \textbf{st\_near} & \textbf{Avg} \\
\midrule
\group{12}{3D Gaussian Splatting}
MonoGS & PSNR & \textbf{23.44} & \textbf{24.66} & \textbf{24.85} & \underline{20.60} & \textbf{23.98} & \textbf{19.35} & \underline{20.63} & \textbf{28.36} & \textbf{27.15} & \textbf{23.67} \\
 & SSIM & \textbf{.782} & \underline{.794} & \textbf{.838} & \underline{.711} & \underline{.805} & \underline{.673} & .712 & \textbf{.896} & \textbf{.874} & \textbf{.787} \\
 & LPIPS & \underline{.247} & .219 & \underline{.211} & \underline{.338} & \underline{.209} & .416 & .330 & \underline{.103} & \underline{.164} & \underline{.249} \\
GS-ICP SLAM & PSNR & 18.14 & \underline{23.61} & 20.95 & 16.13 & 19.71 & \underline{17.75} & 17.87 & 21.76 & 22.38 & 19.81 \\
 & SSIM & .713 & \textbf{.837} & .764 & .685 & .741 & \textbf{.725} & .720 & .736 & .786 & .745 \\
 & LPIPS & .289 & \textbf{.138} & .220 & .346 & .246 & \textbf{.356} & \underline{.299} & .210 & .236 & .260 \\
Photo-SLAM & PSNR & \underline{20.97} & 21.98 & \underline{22.76} & \textbf{21.08} & \underline{23.72} & 16.52 & 20.19 & \underline{27.23} & \underline{25.72} & \underline{22.24} \\
 & SSIM & \underline{.745} & .761 & \underline{.780} & \textbf{.760} & \textbf{.824} & .638 & \underline{.731} & \underline{.888} & \underline{.859} & \underline{.776} \\
 & LPIPS & \textbf{.233} & \underline{.175} & \textbf{.153} & \textbf{.243} & \textbf{.160} & \underline{.408} & \textbf{.266} & \textbf{.091} & \textbf{.138} & \textbf{.207} \\
\midrule
\group{12}{Triangle Splatting}
TS-SLAM & PSNR & 19.36 & 16.32 & 18.71 & 19.57 & 21.31 & 15.51 & \textbf{20.90} & 20.84 & 21.70 & 19.36 \\
 & SSIM & .662 & .721 & .738 & .661 & .694 & .622 & \textbf{.744} & .840 & .830 & .724 \\
 & LPIPS & .369 & .309 & .309 & .383 & .268 & .485 & .361 & .156 & .196 & .315 \\
\ours & PSNR & 18.68 & 21.98 & 20.24 & 17.22 & 21.62 & 15.28 & 19.40 & 23.33 & 22.09 & 19.98 \\
 & SSIM & .675 & .758 & .747 & .652 & .733 & .641 & .708 & .818 & .809 & .727 \\
 & LPIPS & .370 & .257 & .295 & .421 & .247 & .490 & .339 & .186 & .234 & .315 \\
\bottomrule
\end{tabular}}\\[2pt]
\end{table}

\begin{table}[h]
\centering\caption{Rendering quality on ScanNet.}
\label{tab:renderscn}
\footnotesize\setlength{\tabcolsep}{2.5pt}
\resizebox{\columnwidth}{!}{%
\begin{tabular}{llccccccc}
\toprule
\textbf{Method} & & \textbf{0000} & \textbf{0059} & \textbf{0106} & \textbf{0169} & \textbf{0181} & \textbf{0207} & \textbf{Avg} \\
\midrule
\group{9}{3D Gaussian Splatting}
MonoGS & PSNR & \underline{19.82} & \underline{19.57} & \underline{20.42} & \underline{21.96} & \underline{21.33} & \underline{21.13} & \underline{20.71} \\
 & SSIM & .745 & .748 & \underline{.815} & \underline{.818} & \underline{.836} & \underline{.800} & \underline{.794} \\
 & LPIPS & .464 & .392 & .332 & .317 & \underline{.399} & .393 & .383 \\
GS-ICP SLAM & PSNR & \textbf{21.69} & \textbf{19.84} & \textbf{23.76} & \textbf{25.70} & \textbf{22.87} & \textbf{24.51} & \textbf{23.06} \\
 & SSIM & \textbf{.785} & \underline{.770} & \textbf{.871} & \textbf{.856} & \textbf{.867} & \textbf{.856} & \textbf{.834} \\
 & LPIPS & \textbf{.279} & \textbf{.279} & \textbf{.160} & \textbf{.178} & \textbf{.227} & \textbf{.182} & \textbf{.218} \\
Photo-SLAM & PSNR & 18.57 & 18.69 & 18.14 & 19.60 & 13.45 & 18.66 & 17.85 \\
 & SSIM & \underline{.760} & \textbf{.779} & .771 & .813 & .709 & .767 & .767 \\
 & LPIPS & \underline{.302} & \underline{.286} & \underline{.306} & \underline{.295} & .514 & \underline{.336} & \underline{.340} \\
\midrule
\group{9}{Triangle Splatting}
TS-SLAM & PSNR & 17.54 & 18.99 & 16.82 & 18.61 & 13.96 & 18.91 & 17.47 \\
 & SSIM & .683 & .733 & .722 & .742 & .730 & .763 & .729 \\
 & LPIPS & .480 & .366 & .408 & .438 & .472 & .418 & .430 \\
\ours & PSNR & 16.99 & 16.99 & 16.77 & 17.43 & 17.23 & 19.57 & 17.50 \\
 & SSIM & .673 & .672 & .704 & .698 & .768 & .769 & .714 \\
 & LPIPS & .511 & .461 & .452 & .507 & .447 & .410 & .465 \\
\bottomrule
\end{tabular}}
\end{table}

\begin{figure}[h]
\centering\includegraphics[width=\columnwidth]{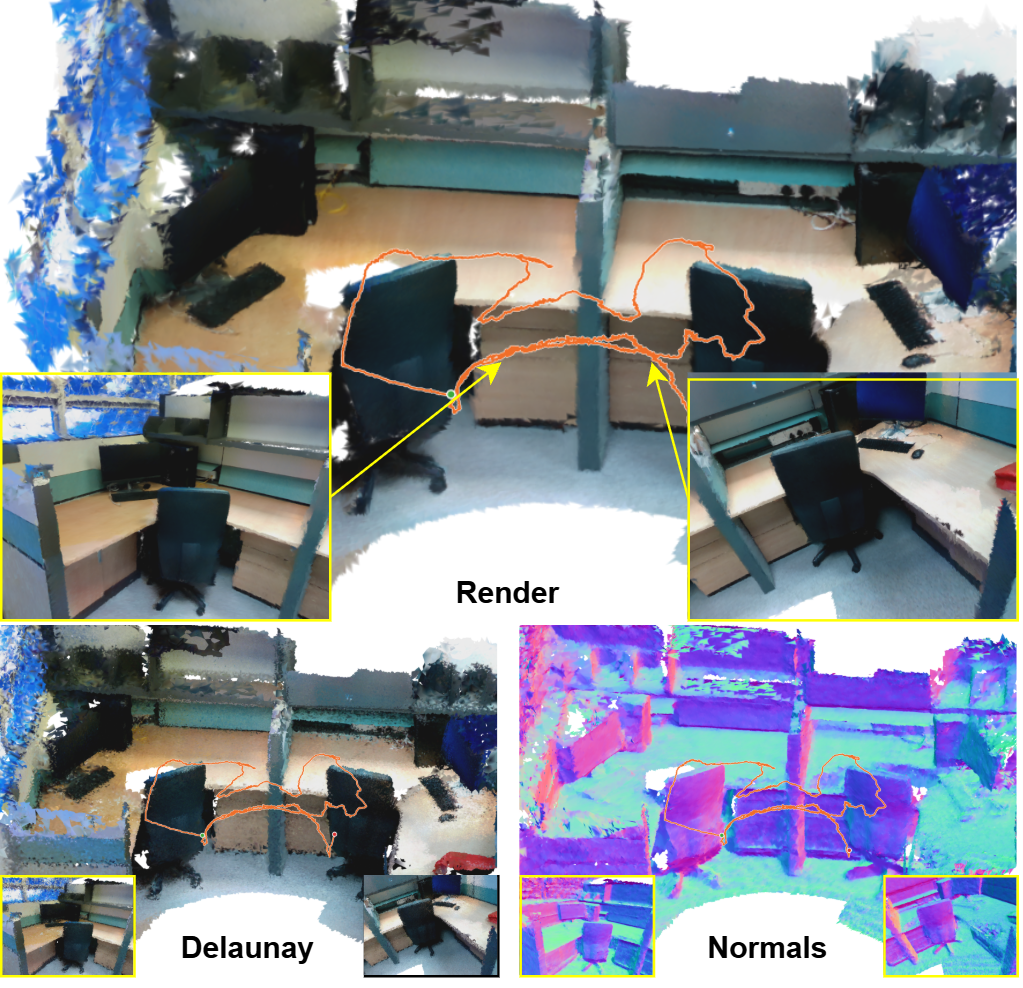}
\caption{\textbf{Qualitative real-world experiment.} Depicted are a set of cubicle cells in a typical office environment, which we mapped on-device using a Jetson AGX Orin and handheld D435 RGB-D camera. From this map we extract the colored Delaunay mesh (bottom left). The render is shown in the top panel, and the corresponding surface normals are shown in the bottom right.}
\label{fig:real-world}
\end{figure}

\subsection{Real-World Experiments}
We conduct qualitative real-world experiments to test our method's performance under real-world conditions and sensor noise. Besides the desk-scale example shown in Fig.~\ref{fig:hook}, we also validate it on a set of office cubicles in Fig.~\ref{fig:real-world}, where a handheld camera was moved around to capture the perpendicular desk structure. The mapping was done entirely online on a Jetson AGX Orin (pictured in Fig.~\ref{fig:hook}). The resulting surface normals and colored Delaunay mesh demonstrates the capability of our proposed method in generating photorealistic meshes.

\subsection{Resource Usage and Runtime}
Our method has comparable GPU memory usage during runtime as TS-SLAM as shown in Table~\ref{tab:footprint}, while operating at sensor rate on the Jetson AGX Orin. We contrast this with existing 3DGS SLAM methods, which tend to use substantially more GPU memory, for both photometric and ICP-based tracking methods alike. Photo-SLAM is the only method we tested that natively runs on the Jetson at an acceptable FPS, yet as shown in Fig.~\ref{fig:geom} its map geometry is not accurate for high quality mesh generation. We compare as an average over the TUM RGB-D sequences reported in Table~\ref{tab:tum}. `X' means the method cannot run on the Jetson either by resource constraint or without substantial changes to the released implementation.

\begin{table}[t]
\centering\caption{Resource consumption comparison on TUM RGB-D.}
\label{tab:footprint}
\scriptsize\setlength{\tabcolsep}{2pt}
\begin{tabular}{lcccc}
\toprule
Method & GPU (GB) & \# primitives & map (MB) & Jetson FPS \\
\midrule
\group{5}{3D Gaussian Splatting}
Photo-SLAM & 0.48--1.44 & 44--78\,k & 12--20 & 23 \\
GS-ICP SLAM & 2.2--6.8 & 0.5--2.0\,M & 35--133 & X \\
MonoGS & 1.5--2.5 & 30--41\,k & \textbf{1.9}--\textbf{2.7} & X \\
\midrule
\group{5}{Triangle Splatting}
TS-SLAM & 0.47--0.53 & \textbf{24}--\textbf{31}\,k & 2.5--3.2 & X \\
\ours & \textbf{0.39}--\textbf{0.52} & \textbf{24}--58\,k & 3.4--8.1 & \textbf{36} \\
\bottomrule
\end{tabular}\\
\end{table}

\subsection{Ablation Study}
\label{subsec:ablation}
We ablate our two coupled contributions. We first show that standard point-to-point or point-to-plane registration is insufficient when applied to triangle primitives, and that our method described improves registration accuracy. We then show that our translation-restricted mapper is necessary for accurate tracking and geometry.

In Table~\ref{tab:metric}, we compare our use of GICP over the local triangle soup, with point-to-point and point-to-plane (using triangle normals) based registration. Our findings are that in general, point-wise costs and the triangles’ own normals are unreliable; a large number of sequences have runs that are bi-modally distributed in terms of ATE. Note that our ablation in this table presumes the use of our mapper-side component in Table~\ref{tab:pin}, without which the tracking degrades.

\begin{table}[h]
\centering\caption{Ablation on the registration metric. We measure ATE RMSE (cm), PSNR (dB) and rendered depth L1 (cm), averaged over the sequences of each dataset.}
\label{tab:metric}
\footnotesize\setlength{\tabcolsep}{3pt}
\resizebox{\columnwidth}{!}{%
\begin{tabular}{lccccccccc}
\toprule
 & \multicolumn{3}{c}{TUM RGB-D} & \multicolumn{3}{c}{Replica} & \multicolumn{3}{c}{ScanNet} \\
\cmidrule(lr){2-4}\cmidrule(lr){5-7}\cmidrule(lr){8-10}
Registration & ATE & PSNR & L1 & ATE & PSNR & L1 & ATE & PSNR & L1 \\
\midrule
Point-to-point & 4.50 & 19.13 & 6.57 & 4.37 & 29.91 & 1.81 & 27.1 & 16.96 & 17.0 \\
Point-to-plane & 8.49 & 19.67 & 6.39 & 0.28 & 33.00 & 0.68 & 27.8 & 17.09 & 15.6 \\
GICP (\ours) & \textbf{1.76} & \textbf{20.30} & \textbf{5.48} & \textbf{0.17} & {33.02} & \textbf{0.67} & \textbf{19.5} & \textbf{17.50} & \textbf{13.0} \\
\bottomrule
\end{tabular}}
\end{table}

In Table~\ref{tab:pin}, we denote with `free' the standard triangle splatting optimizer where the triangle vertices are free to move, and ours where the optimization is constrained to be non-translating. For completeness, we also report `re-centered', where the mapper optimizes the vertices freely and then re-centres each triangle on its tracker-defined centroid; this is marginally worse, and is technically incorrect since Adam's moment estimates still accumulate the discarded translational gradient. Our method also yields more consistent surface orientation on all Replica sequences by a marginal amount ($0.5-1^\circ$).

\begin{table}[h]
\centering\caption{Ablation on the mapper's optimization on triangle positions. We measure ATE RMSE (cm), PSNR (dB) and rendered depth L1 (cm), averaged over the sequences of each dataset.}
\label{tab:pin}
\footnotesize\setlength{\tabcolsep}{3pt}
\resizebox{\columnwidth}{!}{%
\begin{tabular}{lccccccccc}
\toprule
 & \multicolumn{3}{c}{TUM RGB-D} & \multicolumn{3}{c}{Replica} & \multicolumn{3}{c}{ScanNet} \\
\cmidrule(lr){2-4}\cmidrule(lr){5-7}\cmidrule(lr){8-10}
Means & ATE & PSNR & L1 & ATE & PSNR & L1 & ATE & PSNR & L1 \\
\midrule
Free  & 5.11 & 20.07 & 6.71 & 0.23 & \textbf{33.28} & 0.70 & 22.9 & {17.39} & 14.4 \\
Fixed, re-centered & 1.87 & 19.98 & 5.71 & \textbf{0.17} & 32.48 & 0.75 & 21.9 & 17.29 & 13.5 \\
Subspace (\ours) & \textbf{1.76} & \textbf{20.30} & \textbf{5.48} & \textbf{0.17} & 33.02 & \textbf{0.67} & \textbf{19.5} & \textbf{17.50} & \textbf{13.0} \\
\bottomrule
\end{tabular}}
\end{table}

\section{Limitations}
A limitation of our method is primarily lower mapping quality than 3DGS based methods. This is partly a result of design decisions for computational efficiency, such as using 0-th degree SHs or more conservative densification behaviors, and partly due to the relatively new triangle splatting representation for SLAM, compared to 3DGS which has been optimized for some time. Future works may consider optimization techniques that may improve mapping quality while maintaining the spatial robustness of the primitives for geometric tracking algorithms.

\section{Conclusion}
In this work, we present a real-time triangle splatting SLAM system, which enables on-device localization and mapping with both photorealistic renders, and favorable geometric properties to support embodied AI tasks. Our system does this while maintaining competitive tracking at more than 100 FPS, or at sensor rate on a Jetson embedded platform.

% --- BIBLIOGRAPHY SETTINGS ---
\bibliographystyle{IEEEtran} % Use the official IEEE style
\bibliography{references}    % This looks for a file named references.bib

\newpage

\vfill

\end{document}